\documentclass[letterpaper]{article} 
\usepackage[]{aaai25}  
\usepackage{times}  
\usepackage{helvet}  
\usepackage{courier}  
\usepackage[hyphens]{url}  
\usepackage{graphicx} 
\usepackage{natbib}  
\usepackage{caption} 
\usepackage{tikz}
\usepackage{multirow}

\usepackage{algorithm}
\usepackage{algorithmic}

\usepackage{newfloat}
\usepackage{listings}
\DeclareCaptionStyle{ruled}{labelfont=normalfont,labelsep=colon,strut=off} 
\floatstyle{ruled}
\newfloat{listing}{tb}{lst}{}
\floatname{listing}{Listing}
\title{Infusing Prompts with  Syntax and Semantics}

\author{
    Anton Bulle Labate,
    Fabio Gagliardi Cozman
}
\affiliations{
    Universidade de São Paulo \\
    antonlabate@usp.br,
    fgcozman@usp.br
   
}

\begin{document}

\maketitle

\begin{abstract}
Despite impressive success, language models often generate outputs with flawed linguistic structure. We analyze the effect of directly infusing various kinds of syntactic and semantic information into large language models. To demonstrate the value of our proposals, we focus on the translation of natural language queries to SQL, in particular dealing with languages with less resources than English, to better investigate how much help we can get from low cost syntactic and semantic information. We show that linguistic analysis can significantly boost language models, to the point that we have surpassed previous best systems. 
\end{abstract}

\section{Introduction}

In this paper we introduce techniques that 
resort to syntactic and semantic information so
as to improve the performance of language models.
Intuitively, one would expect that solid linguistic
knowledge about human language, and  honed linguistic
tools, should be useful in natural language processing.
Indeed, one must hope that suitable syntactic and semantic
analysis should be useful in preventing a number of 
failure modes that have been observed even in large language
models \cite{leivada2022dalle,yao2022structural}.

Here we explore this intuition, and we indeed find that
syntactic and semantic information are rather valuable, 
at least in settings where existing data-driven resources 
are limited by practical or economic circumstances.
In fact, we show that relatively simple ideas
lead to key gains in performance. In essence, we infuse 
syntactic and semantic information into textual prompts, 
instead of changing computational paths as investigated 
in previous proposals. In doing so we add another 
``prompting pattern'' to the library of patterns that have
been explored in recent years.

To be concrete, we focus on the translation
of natural language questions to SQL queries; 
our proposals should be easily transferable to other 
tasks in natural language processing. 
We focus on languages other than English, where the
quality of query translation has evolved at a lower
pace than it has for English,
so as to face more challenging scenarios
that can shed light the value of linguistic devices.  
We show that, at a relatively low cost, one can significantly boost the performance of language models --- for instance, we have been able to  surpass the previous state of the art result on the versions of the Spider benchmark for Portuguese and French queries.
We hope these results will drive people to revisit syntactic and semantic information that can have as much effect as large amounts of data while consuming less resources. 
In fact, there seems to be a trade-off: 
one can make up for limited data/computing resources by
resorting to linguistic tools, and conversely one may 
dim the value of syntactic and semantics analysis by 
resorting to ever increasing data resources --- one might
conjecture that a {\em really} large language model accrues
less benefit from linguistic analysis as it somehow
``learns'' or absorbs the relevant bits of analysis from data.

The paper is organized as follows.
In Section~\ref{section:Background} we briefly go through relevant 
previous work;
in Section \ref{section:Proposal} we describe our proposals, and
in Section~\ref{section:exp} we describe our experiments.
Section~\ref{section:Conclusion} wraps up with some discussion and conclusions.

\section{Background and Related Work}\label{section:Background}

There are a number of techniques that aim at infusing morphological/syntactic/semantic information into language models (LMs).
\citeauthor{Punyakanok2008} (\citeyear{Punyakanok2008}) is an early 
example of syntactic infusion in role labeling (without transformers). 
Other   efforts  mix classic linguistic
results within learning techniques \cite{sogaard-goldberg-2016-deep}.
%
\citeauthor{Shiv2019} (\citeyear{Shiv2019}) use a positional
embedding that exploits relations in a syntax tree, so as to get some of
the syntactic structure into an LM. Clearly this requires changes to the
original transformer model. Another effort, by 
\citeauthor{Qian2021} (\citeyear{Qian2021}), modifies the training of
the LM by asking the model to predict not only words but also parse trees.

Morphological parse trees have also been used during training;
\citeauthor{selfattentionBai2021} (\citeyear{selfattentionBai2021}) have
self-attention modules for each relation from a morphological tree.
And morphological trees have  been mapped to vectors by embeddings \cite{zanzotto-etal-2020-kermit},
with tests on encoder-only models for classification tasks.
\citeauthor{currey2019incorporating} (\citeyear{currey2019incorporating}) enhance
translation performance  by infusing morphology trees as input to trained encoders, and also improve translation with encoder-decoder models by pre-training the model on a parsing task. 
\citeauthor{sg-net} (\citeyear{sg-net}) uses a network to insert morphological relations
into the self-attention mechanism of a LM.
Regarding the transformer's attention mechanism, \cite{9303437} uses the information of the words related to each other in  the attention mechanism.

Structures beyond morphological relations have also been infused
into LMs. For instance, \citeauthor{bert-parse} (\citeyear{bert-parse}) use a graph similar to
the Abstract Meaning Representation (AMR) of the input,  during training
for an encoder architecture (with experiments in various classification tasks). 
Another work that relies on syntactic relations, by \citeauthor{nmt} (\citeyear{nmt}),
introduces  embeddings for syntactic features and then combine those embeddings, and positional
encodings. Note that  we also
infuse syntactic information but we rely on a single
embedding and do not make changes to positional
encodings nor to   architectural elements of the language model.

We now add a few comments about the (rather large) literature
on natural language to SQL translation, referred to as NL2SQL. 
Early efforts in NL2SQL were
rule-based \cite{nalir,athena}, but now the best systems rely on deep networks
trained with large datasets such as WikiSQL \cite{zhong2017seq2sql} and
Spider \cite{yu2019spider}. However, transformers still struggle with
phrasal structures, for instance ones that carry user intentions with multiple particular restrictions (that is, different adjectives for instance modifying different nouns).
These difficulties are particularly acute in languages other than
English for which datasets with examples are small. An alternative to this problem could be the use of a translation module for the user question to English, and then
resort to NL2SQL in English.   Nevertheless, this approach would bind the user   to an external translation API or to a translation model from a given language to English,  burdening the overall translation. In addition, \cite{jose2021mrat} and \cite{jose2023multilingual} have found that a single multilingual transformer 
works better than a translation to English (followed by a  
English-trained transformer), for a number of source languages. 

Efforts as reported in \cite{dou2023multispider}, \cite{jose2023multilingual}, \cite{min-etal-2019-pilot}, \citet{bakshandaeva2022pauq} and \citet{almohaimeed2024ar} have developed datasets in French, Spanish, Portuguese, Chinese, Russian, Arabic and other languages for the task of translation of Natural Language to SQL. 

A Chinese version Spider, CSpider, was one of the first translations of the Spider dataset. \citeauthor{min-etal-2019-pilot} manually translated and verified the questions in English, 
following the most natural ways to adapt sentences to the Chinese language and cultural context. 

In \cite{bakshandaeva2022pauq}, the authors present the first translation of Spider to Russian. In that work, besides manually translating the question, as in CSpider, the authors augmented the dataset questions, by providing samples of types of questions regarding columns with binary values, columns containing date and time values, and the ones that have a fuzzy and partial
match with the database content. 

In \cite{jose2023multilingual}, the authors translated the Spider dataset to French, Portuguese and Spanish. However, as in \cite{almohaimeed2024ar}, in the development of the Arabic version of Spider, they use automatic translation for  the work. Nonetheless, the authors of both works manually verified samples from the translation, with the latter also using an evaluation from a LLM for the translation to ensure its quality.

Several of these previous efforts used translated   training data and found strong baselines that trained models with these datasets achieve, even though smaller, but yet comparable results to the ones obtained through training with English data only \cite{min-etal-2019-pilot}. This shows the feasibility of training models directly using the target language as input to the language models, allowing one to unlock the many benefits of using an independent model for the translation from an input query in his native language to SQL.

There have been also efforts that employ parsers within NL2SQL. For instance, \citeauthor{Yaghmazadeh2017}(\citeyear{Yaghmazadeh2017}),\citeauthor{iyer2017learning} (\citeyear{iyer2017learning}) parse the input and map it to a representation near, such as the SQL skeleton, or even the proper SQL query. Therefore neither of them actually applies linguistic analysis of the phrase structure. On the other hand, \citet{nalir}, even though relying on a framework that requires human feedback for the generated queries, makes use of the morphological tree of the phrase as an intermediate representation of the user query to generate by rule-mapping the query, not using transformers. Neither of these approaches infuse formal phrase structure information into transformers, nor  use  the input syntax or semantics, as we do in our method. 

\begin{figure*}[t]
\centering
\begin{tikzpicture}
\node[fill=orange!20,rectangle,rounded corners]
(inputtext) at (5.2,2.9) 
{\begin{tabular}{l}
Inscrivez l'année \\
de création, le nom \\ 
et le budget de chaque \\
département.
\end{tabular}};
\node (input) at (4,4) {\bf Input};
\node[fill=black!20,rectangle,draw] (parser) at (6,6)
{\bf Parser};
\node (sum) at (8,4) {\Huge $\bigoplus$};
\draw[very thick,->] (input) -- (4,6) -- (parser);
\draw[very thick,->] (input) -- (sum);
\draw[very thick,->] (parser) -- (8,6) -- (sum);
\node[fill=blue!20,rectangle,rounded corners]
(parsed) at (9.9,6) 
{\small \begin{tabular}{l}
[row] year; \\
dobj [row] name; \\
conj [row] budget; \\
conj [row] department; \\
pobj
\end{tabular}};
\node[fill=black!20,rectangle,draw] (roberta) at (14,4)
{\bf RoBERTa};
\draw[very thick,->] (sum) -- (roberta);
\node[fill=blue!20,rectangle,rounded corners]
(robertatext) at (10.6,2.35) 
{\small \begin{tabular}{l}
Inscrivez l'année \\
de création, le nom \\ 
et le budget de chaque \\
département.
[row] year; \\
dobj [row] name; \\
conj [row] budget; \\
conj [row] department; \\
pobj
\end{tabular}};
\node[fill=black!20,rectangle,draw] (llm) at (14,0)
{\bf LLM};
\draw[very thick,->] (roberta) -- (llm);
\node[fill=blue!20,rectangle,rounded corners]
(llmtext) at (17.8,4) 
{\small \begin{tabular}{l}
Inscrivez l'année de création, \\
le nom et le budget de chaque \\
département. [row] year; \\
dobj [row] name; conj [row] budget; \\
conj [row] department; pobj $|$ \\
department : department.name , \\
department.creation , \\
department.budget\_in\_billions , \\
department.department\_id , \\
department.num\_employees $|$ \\
head : head.name , head.age , \\
head.head\_id , \\
head.born\_state $|$ management : \\
management.department\_id , \\
management.temporary\_acting , \\
management.head\_id $|$ \\
management.head\_id = head.head\_id $|$ \\
management.department\_id = \\
department.department\_id
\end{tabular}};
\node (output) at (9,0) {\bf Output};
\draw[very thick,->] (llm) -- (output);
\node[fill=orange!20,rectangle,rounded corners]
(outputtext) at (9,-0.7)
{\begin{tabular}{l}
select \_ from \_ $|$ select creation , name , \\ budget\_in\_billions from department
\end{tabular}};
\end{tikzpicture}
\caption{The pipeline, from input to output.}
\label{figure:Pipeline}
\end{figure*}
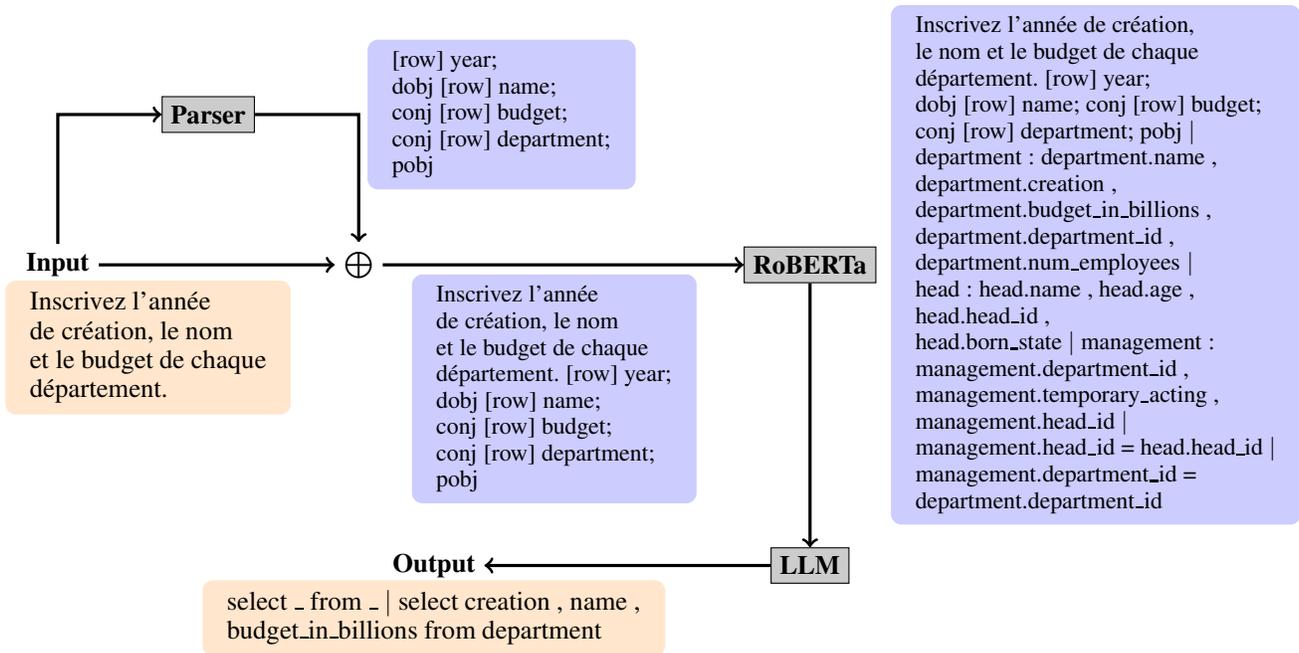

\begin{table*}\small
  \centering
  \renewcommand{\arraystretch}{1.2}
  \begin{tabular}{|c|c|c|c|c|c|c|c|c|c|c|c|c|}
    \hline
    {\multirow{2}{*}{\textbf{Model}}} & \multicolumn{4}{c|}{\textbf{French}} &  \multicolumn{4}{c|}{\textbf{Spanish}}& \multicolumn{4}{c|}{\textbf{Portuguese}}\\
    \cline{2-13}
    & \textbf{EM} & \textbf{EX}  & \textbf{vEM} & \textbf{vEX} & \textbf{EM} & \textbf{EX} & \textbf{vEM} & \textbf{vEX} &\textbf{EM}  &\textbf{EX}& \textbf{vEM} & \textbf{vEX}\\
    \hline
    T5 without info  & 0.612 & 0.618  & - & -  & 0.582 & 0.612  & - & -  &  0.550 & 0.575  & - & - \\ \hline
    T5 with syntax & 0.623  & 0.649 & 1.8 & 5.0 &  0.616 & 0.631 & 5.8 & 3.1 & 0.591 & 0.625  & 7.4 & 7.6   \\
      \hline
    T5 with AMR  & 0.640 & 0.693 & 4.6  & 12.1 & 0.635 & 0.691 & 9.14  & 12.9 & 0.639 & 0.682 & 16.2 & 18.5 \\ 
      \hline
      T5 with syntax + AMR  & 0.638 & 0.684 & 4.25  & 10.68 & 0.639  & 0.679 & 9.79  & 10.95 & 0.625  & 0.663 & 13.64 & 15.3 \\ 
      \hline
      Bart without info & 0.199 & 0.199 & - & - & 0.199 & 0.199 & - & - & 0.182  & 0.181 & - & -  \\ \hline
      Bart with syntax & 0.234 & 0.249 &  17.6 & 25.0 & 0.223 & 0.236 & 12.1  & 18.6 & 0.216 & 0.221   &  40.2 & 37.3 \\
      \hline
      Bart with AMR & 0.215 & 0.225 & 8.0 & 13.0 & 0.242 & 0.262 & 21.6 & 31.7 & 0.222  & 0.245  & 40.2 & 52.4  \\
    \hline
    Bart with syntax + AMR & 0.269  & 0.286 & 35.18 & 43.72 & 0.255 & 0.282 & 28.14 & 41.71 & 0.287  &  0.311 & 57.69 & 40.72  \\
    \hline
  \end{tabular} 
  

\vspace*{1ex}

  \centering
  \renewcommand{\arraystretch}{1.2}
  \begin{tabular}{|c|c|c|c|c|c|c|c|c|c|c|c|c|}
    \hline
    {\multirow{2}{*}{\textbf{Model}}} & \multicolumn{4}{c|}{\textbf{French}} &  \multicolumn{4}{c|}{\textbf{Spanish}}& \multicolumn{4}{c|}{\textbf{Portuguese}}\\
    \cline{2-13}
    & \textbf{EM} & \textbf{EX}  & \textbf{vEM} & \textbf{vEX} & \textbf{EM} & \textbf{EX} & \textbf{vEM} & \textbf{vEX} &\textbf{EM}  &\textbf{EX}& \textbf{vEM} & \textbf{vEX}\\
    \hline
    T5 without info  & 0.618  & 0.632  & - & -  & 0.591  & 0.618  & - & - &  0.586 & 0.602 & - & - \\ \hline
    T5 with syntax & 0.657  & 0.673 & 6.3 & 6.5 &  0.631 & 0.636 & 6.8 & 2.9  & 0.630  & 0.649  & 7.4 & 7.9  \\
      \hline
    T5 with AMR  & 0.669 & 0.700 &  8.2 & 10.8  & 0.653  & 0.687  & 10.5 & 11.2 & 0.653  &   0.687  & 11.4 & 14.1 \\ 
      \hline
      T5 with syntax + AMR  & 0.652 & 0.695 & 5.5  & 9.97 & 0.651 & 0.708 & 10.15  & 14.56 &  0.649  & 0.697 & 10.75 & 15.78 \\ 
      \hline
      Bart without info & 0.184 & 0.184 & - & - & 0.197 & 0.199 & - & - & 0.163  & 0.171 & - & -  \\ \hline
      Bart with syntax &  0.238 & 0.240  & 29.3  & 30.4  & 0.218  & 0.220  & 10.7  & 10.6 & 0.218 & 0.222   &  33.1 & 29.9 \\
      \hline
      Bart with AMR & 0.227  & 0.240 & 23.4 & 30.4 & 0.210 & 0.224    & 6.6  & 12.6 & 0.189 & 0.196 & 15.9  & 14.6 \\
    \hline
    Bart with syntax + AMR & 0.279 & 0.299 & 51.63 & 62.5 & 0.276 & 0.289 & 40.1 & 45.23 &  0.285 & 0.305  & 74.85 &  78.36 \\
    \hline

  \end{tabular}
\caption{Model performance after training for 32 (top) and 128 (bottom) epochs.}  \label{table:theTable}
\end{table*}

\section{Infusing Syntactic and Semantic Information}\label{section:Proposal}

In tasks sensitive to input phrases, that is, tasks in which a little modification in the input phrase can produce a perceptive difference in the output, it is important for the model to capture and understand the relations between the terms of the phrase as to generate the correspondent output, reflecting them, and consequently, reflecting the user's intention that they convey as a whole.
For instance, in the task of translation from natural language to natural language, adjectives are supposed to only modify corresponding nouns, and such connections must be respected in translated output. However, transformers not always are capable of interpreting correctly these relations, resulting in outputs that flaw, or do not translate or represent the relations required by the user's phrase \cite{leivada2022dalle}.
The same is observable in the task of translating natural language requests to SQL queries, in which the model must identify correctly to which term a particular restriction refers to in order to generate the correct SQL answer, for instance. 

Intuitively, one might think that syntactic information is bound
to improve LMs; however, it is not easy to see how to infuse
such information in ways that actually make a positive difference 
at low cost.
We explore a direct strategy where syntactic information is 
  fed as part of the LM prompt. To do so, we 
concatenate syntactic information with the raw input text. 
We use such enlarged input  during
training and inference, even though it is possible to produce
outputs without enlarging the input (with syntactic and semantic information)
 during inference.
Thus we do not have to change the architecture of the LM of choice.

The resulting pipeline is depicted in Figure \ref{figure:Pipeline}:
the input goes through syntactic and semantic analysis, and then it
is concatenated with the result of that analysis; the concatenated
text along with the schema related to the request goes to the LM.


We rely on two types of information about the phrase: dependency trees \cite{jurafsky}, representing the syntax, and AMR data, for the semantics. The \textit{Abstract Meaning Representation} (AMR) is a semantic representation framework that captures the meaning of a sentence 
through a directed acyclic graph where nodes correspond to concepts 
or entities, and edges represent various  relationships 
between them. Concepts are typically abstract and 
language-independent in AMR, allowing for a higher level of semantic 
generalization. The relationships between concepts capture roles, 
attributes, and dependencies.

For instance, as a running example, take an input sentence  as follows for a translation from French to SQL: 
\begin{quote}
Inscrivez l'année de création, le nom et le budget de chaque département.\footnote{In English: "List the creation year, name and budget of each department."}
\end{quote}
The text produced out of the dependency tree for this input is:

\begin{quote} \small
Inscrivez ROOT Inscrivez PROPN [année, .]\\
l' det année NOUN []\\
année obj Inscrivez PROPN [l', création]\\
de case création NOUN []\\
création nmod année NOUN [de, ,, nom]\\
, punct création NOUN []\\
le det nom NOUN [] \\
nom appos création NOUN [le, budget] \\
et cc budget NOUN [] \\
le det budget NOUN [] \\
budget conj nom NOUN [et, le, département] \\
de case département NOUN [] \\
chaque det département NOUN [] \\
département nmod budget NOUN [de, chaque] \\
. punct Inscrivez PROPN [] \\
\end{quote}

In this representation, the syntactic function in the phrase is written for each term of the phrase, followed by the term to which it refers to (to which it is dependent to in the relation, its ``syntactic head"), its Part of Speech tag (morphology) and the terms that depends on it (``syntactic dependents").

However, relations such as determinants, that do not significantly alter the phrase's meaning, are also included in this representation. Because we are only interested in teaching the model the core syntactic relations of the phrase (the ones that truly carry its meaning), we only provide the model with a particular group of those relations: subjects, objects and conjunctions. 
 
To translate a dependency tree into running text, we follow a proposal by \citeauthor{Li2021} (\citeyear{Li2021}): for each of these particular relations, we initiate the sentence with [row], followed by the term that is the head of the syntactic function and and the name of the syntactic function, separated  by ``;''. For this running example, the prompt for the language model, following this method, is:

\begin{quote} \small
Inscrivez l'année de création, le nom et le budget de chaque département. [row] year; dobj [row] name; conj [row] budget; conj [row] department; pobj    
\end{quote}
  
To infuse semantic information, we turn  AMR graphs to  text by  removing spaces and relying on AMR brackets  
to separate the parser's output, in order to teach the model how to make the correct associations in   PropBank notation \cite{10.1162/0891201053630264}. 
For instance, if we have the same input sentence as above, the running text
produced out of the AMR analysis is:
\begin{quote} \small
    (l / list - 01 :ARG1 (a / and :op1 (y / year :time - of (c / create - 01 :ARG1 (d / department :mod (e / each)))) :op2 (n / name :poss d) :op3 (b / budget :poss d)))
\end{quote}
When combining it with the target sentence, we have:
\begin{quote} \small
    Inscrivez l'année de création, le nom et le budget de chaque département. (l / list - 01 :ARG1 (a / and :op1 (y / year :time - of (c / create - 01 :ARG1 (d / department :mod (e / each)))) :op2 (n / name :poss d) :op3 (b / budget :poss d)))
\end{quote}
Finally, we have explored the impact of different ways of prompting   to the model, through syntax and semantics, in independent experiments, but we also thought it would be interesting to see how the model would behave with the interaction between these two types of information about the phrase. Thus, we also presented to the models phrases that combine the two formats above. In order to separate the semantic representation from the syntactic structure, as we did to separate the syntactic terms, we include before the linearized AMR graph the tag "[AMR]". For the same input sentence, the resulting text combining syntax and semantics is:
\begin{quote} \small
    Inscrivez l'année de création, le nom et le budget de chaque département. [row] year; dobj [row] name; conj [row] budget; conj [row] department; pobj [AMR] (l / list - 01 :ARG1 (a / and :op1 (y / year :time - of (c / create - 01 :ARG1 (d / department :mod (e / each)))) :op2 (n / name :poss d) :op3 (b / budget :poss d)))
\end{quote}

We adopt the RESDSQL framework for NL2SQL \cite{li2022resdsql}. In this framework, a trained encoder model (Roberta \cite{liu2019roberta}), given the user's request, selects the most relevant tables and columns accordingly, and presents only them along the user question to the transformer. Besides, in RESDSQL, the language model used to generate SQL first decodes the SQL structure from the structure, without assigning values, and then the language model generates the complete SQL output. 
It is important to note that we do not make any changes to the framework, only to the tokens that we send as input to it.

\begin{table}\small
  \centering
  \renewcommand{\arraystretch}{1.2}
  \begin{tabular}{|c|c|c|c|c|}
    \hline
    {\multirow{2}{*}{\textbf{Model}}} & \multicolumn{4}{c|}{\textbf{Chinese}} \\
    \cline{2-5}
    & \textbf{EM} & \textbf{EX}  & \textbf{vEM} & \textbf{vEX} \\
    \hline
    T5 without info  & 0.094 & 0.251  & - & -  \\ \hline
     mT5 without info  & 0.522 & 0.678  & - & - \\ \hline
    mT5 with syntax & 0.550  & 0.710 & 5.36 & 4.72 \\
      \hline
    mT5 with AMR  & 0.595 & 0.736 &  13.98 & 8.55\\ 
      \hline
      mT5 with syntax + AMR  & 0.594 & 0.723 & 13.79  &  6.64 \\ 
      
    \hline
  \end{tabular} 

\vspace*{1ex}

  \centering
  \renewcommand{\arraystretch}{1.2}
  \begin{tabular}{|c|c|c|c|c|}
    \hline
    {\multirow{2}{*}{\textbf{Model}}} & \multicolumn{4}{c|}{\textbf{Chinese}} \\
    \cline{2-5}
    & \textbf{EM} & \textbf{EX}  & \textbf{vEM} & \textbf{vEX} \\
    \hline
   
     mT5 without info  & 0.559 & 0.706   & - & - \\ \hline
    mT5 with syntax & 0.582  & 0.739 & 4.11 &  4.67 \\
      \hline
    mT5 with AMR  & 0.597 & 0.737 & 6.8  & 4.39  \\ 
      \hline
      mT5 with syntax + AMR  & 0.625 & 0.747 & 11.81  & 5.81  \\ 
      
    \hline
  \end{tabular} 
\caption{Chinese model performance after training for 32 (top) and 128 (bottom) epochs.}  \label{table:theTablechinese128}
\end{table}  


\begin{table}\small
  \centering
  \renewcommand{\arraystretch}{1.2}
  \begin{tabular}{|c|c|c|c|c|}
    \hline
    {\multirow{2}{*}{\textbf{Model}}} & \multicolumn{4}{c|}{\textbf{English}} \\
    \cline{2-5}
    & \textbf{EM} & \textbf{EX}  & \textbf{vEM} & \textbf{vEX} \\
    \hline
    T5 without info  & 0.717 & 0.770 & - & -  \\ \hline
    T5 with syntax &  0.722 & 0.768 & 0.7 & -0.26 \\
      \hline
    T5 with AMR  & 0.708  & 0.752  &  -1.26 & -2.34\\ 
      \hline
      T5 with syntax + AMR  & 0.700 & 0.756 & -2.37  & -1.82  \\ 
      \hline
      Bart with no info & 0.303 & 0.326  & - & - \\
      \hline
      Bart with syntax & 0.295& 0.313& -2.64& -3.99 \\
      \hline
      Bart with AMR  & 0.310& 0.327& 2.31& 0.31\\
      \hline
      Bart with AMR + syntax &  0.315 & 0.341  & 3.96 & 4.6\\
      
    \hline
  \end{tabular} 

\vspace*{1ex}

  \centering
  \renewcommand{\arraystretch}{1.2}
  \begin{tabular}{|c|c|c|c|c|}
    \hline
    {\multirow{2}{*}{\textbf{Model}}} & \multicolumn{4}{c|}{\textbf{English}} \\
    \cline{2-5}
    & \textbf{EM} & \textbf{EX}  & \textbf{vEM} & \textbf{vEX} \\
    \hline
   
     T5 without info  & 0.717 &  0.779 & - & - \\ \hline
    T5 with syntax & 0.722 & 0.776 & 0.7 &  -0.39 \\
      \hline
    T5 with AMR  & 0.714  & 0.778 & -0.42  & -0.13  \\ 
      \hline
      T5 with syntax + AMR  & 0.719 & 0.769 & 0.28  & -1.28 \\ 
       \hline
      Bart with no info & 0.298 & 0.307& - & - \\
      \hline
      Bart with syntax & 0.290 & 0.293  & -2.68 & -4.56 \\
      \hline
      Bart with AMR  & 0.296 & 0.321 &-0.67 & 4.56 \\
      \hline
      Bart with AMR + syntax &  0.281 & 0.292  & -5.7& -4.89 \\

    \hline
  \end{tabular} 
\caption{English model performance after training for 32 (top) and 128 (bottom) epochs.}  \label{table:theTableenglish128}
\end{table}  

\section{Experiments}\label{section:exp}

Our experiments investigate whether syntactic and semantic information are meaningful in the task of translating natural language to SQL. This task is rather sensitive: a small
mistake in understanding the user intents can lead to seriously incorrect SQL queries. Hence one might expect linguistic analysis to be valuable. 

\subsection{The (translated) Spider Dataset}

Our experiments were conducted with several languages: 
Chinese, French, Portuguese and Spanish.
We can expect existing LLMs to have been exposed to relatively
little data on these latter languages as compared to English.\footnote{For instance, in the most recent
CommonCrawl corpus (https://commoncrawl.org/), 43.79\%
of all texts are in English, while Chinese, Spanish and French
appear respectively with 5.14\%, 4.56\% and 4.22\% of texts,
and Portuguese appears with meager 2.12\% of texts.}

The non-English datasets we used in our experiments were translated versions 
of the Spider corpus \cite{yu2019spider}, which
seems to be the most popular corpus in NL2SQL.
Obviously, the quality of translations affects our experiments; however, it
should be noted that the same translated datasets (for each one of the
languages) were used in the several compared techniques, so that the relative
performance of these techniques is still meaningful. In any case, the
quality of translations is discussed in the remainder of this section. 

The Chinese version of Spider used has been translated by \cite{min-etal-2019-pilot}. This translated dataset was produced with the work of two NLP researchers and one computer science student. 
For each phrase, the translator was asked to read the English sentence as well as its corresponding SQL query, after that he was asked to translate the sentence to Chinese sticking as much as possible to the literal translation. 
For complex phrases, the translator was allowed to rephrase the English question as to be able to make the most natural Chinese translation. To maintain the diversity of the dataset, particular English expressions were matched to their equivalent in the Chinese language. 
 
The French, Portuguese and Spanish versions of the Spider dataset have been produced by \citet{jose2023multilingual}. For each language, the authors resorted to the automatic translation of the English phrases using the Google Translation API,
followed by  manual evaluation and revision. As automatic translation may distort particular and contextual expressions in the source language, we examined the quality of the process: we run a detailed evaluation of the Portuguese translation of Spider, given that Portuguese is the lowest-resource language among the ones we dealt with and therefore is the most likely to present a challenge to the automatic translator (as one can surmise from Footnote 2). The evaluation process is summarized in the Appendix.

\subsection{Empirical Analysis}


As English NL2SQL systems have already received myriad improvements (so that any individual technique is bound to have minimal impact), we worked with other languages so as to better
grasp the absolute value of our proposals. We have focused on French/Spanish/Portuguese/Chinese translation to SQL. 
We employed translated versions of the Spider (under the CC BY-SA 4.0 license) dataset for training and testing. As for time of the translation of these datasets, used for our training, the Spider's test set was not publicly available, we restrict our validation to the development set only and training with its training set.

Our experiments used  two different architectures of sequence-to-sequence models for fine-tuning, in order to verify the generalizability of the method, for French, Portuguese and Spanish: Bart base \cite{lewis2019bart} and T5 base \cite{raffel2023exploring}. We emphasize that we opted in this set of experiments to use the base versions of T5 and Bart, not their multilingual models, in order to simulate the most adverse scenario one might encounter when choosing a model for a task, in which among the best possible models none of them has been trained with the intended language. For instance, mT5 \cite{xue2021mt5massivelymultilingualpretrained} was not trained with a dataset that contains Croatian; therefore, despite mT5 being multilingual, a Croatian user would still face the challenge to accept to use a model that have not seen his native language and intended use language before in its pre-training (as T5 has not seen as much French, Portuguese or Spanish, the languages we use in our experiments). In this way, we can measure the impact of our proposal in  challenging scenarios. 

For the Chinese models, however, this was not possible to use T5. We only used mT5, since the base version of T5, the model with the best results for French, Portuguese and Spanish in our experiments, struggled with the new language (0.094 for exact-set match and 0.251 for execution accuracy, scores prohibitively low to make any safe conclusions around the impact of our method). Nevertheless, we see these results as complementary to the above, as they lead to same conclusions in a scenario where, even though the model recognizes the language, it still benefits from the infusion, proving that the method works for both cases.

Dependency trees and AMR graphs have been generated by the Spacy parser and amrlib,\footnote{https://github.com/bjascob/amrlib} respectively. For fine-tuning the base models, the experiments were conducted on 2 GPUs NVidia GeForce RTX 3090 24 GB. For the specific fine-tuning of a T5 with 3 billion parameters for Portuguese and French a single A100 GPU with 80 GB was used.
All models were trained using the AdaFactor optimizer, with a learning rate of 1e-4, batch size of 8 and gradient descent after two steps. 
Our code and resulting models will be made publicly available in case the present
text is accepted for publication, under the CC BY-SA 4.0 license.

We evaluated our models using  exact-set-match and execution accuracy.
These metrics are rather complementary, as exact-set-match may overlook more false negatives (one query might be the answer, even though not   the golden one), whereas execution accuracy may overlook more false positives (a value might be the one generated by the golden query, nonetheless by another term in the table, not the one referenced by the golden query). 

To understand the impact of syntactic and semantic information of the input in a model performance and its relevance during training,  we also  trained, for French, Portuguese and Spanish, a Bart model with the raw corresponding (translated) version of Spider, a model with the syntactic information, one with the semantic information infused and one with the two types of knowledge combined. 
We then did the exact same set of training runs for the T5 models and mT5 models, in the case of Chinese. Thus we
generated eight distinct language models per natural latin language, and four for the Chinese language.
Using this scheme, we were able to compare   the results for the training without  any added linguistic information (raw dataset), solely with syntax infused, solely  with semantics infused and with the two pieces of information infused. 

We conducted the set of experiments described in the previous paragraph both for 32 and for 128 training epochs. With these two numbers of training epochs, operating with the same data and language model,   we were able to evaluate the  relevance of the information presented to the progress of  the training process itself. 

Table  \ref{table:theTable} (top) conveys the results for the translations for the French, Portuguese and Spanish corpus for 32 epochs, while Table \ref{table:theTable} (bottom) presents the results of the models trained for 128 epochs with French, Spanish and Portuguese, respectively. Table \ref{table:theTablechinese128} presents the results of the models trained with CSpider. The "EM" and "EX" columns contain values of exact-set match and execution accuracy respectively. 
The vEM and vEX columns contain, respectively, the relative variation between the produced model's exact-set match accuracy and the base model's exact-set match accuracy, and the relative variation between the produced model's execution accuracy and the base model's  execution accuracy, in percentages.

The tables show that the infusion of syntactic and semantic information is valuable during training. Models that were fed during training with the input phrase  syntax or semantic structure displayed higher performance than the corresponding models without any linguistic information, regardless of the language model, for every natural language we tried. Also, this can be verified  both in the lower number of epochs and in the higher number of epochs.  In most experiments, the models that received  semantic information had the higher scores. Thus we can see that syntactic and semantic information lead to better scores.

In addition, syntactic and semantic information accelerate training. Even when trained only for 32 epochs, models with syntactic or semantic information achieve higher results than models without any such information but trained with four times more epochs (that is, with 128 epochs). Therefore, the infusion not only attains better performance within a given number of training epochs, but it also attains a given performance with a smaller number of epochs. 

Particularly, we were able to surpass the state of the art for Portuguese and French when training the models using the T5 architecture with 3 billion parameters and Nat SQL \cite{gan2021natural} as intermediate representation of the user's input and the corresponding SQL query (note that state of art results in the literature did not report  execution accuracy). 
In Table \ref{tablecomparison} we present our best results and compare them with previous reported results (French and Portuguese by \cite{jose2023multilingual}).  
 \begin{table}[t]
\centering
\begin{tabular}{|l|l|l|l|l|}
    \hline
    {\multirow{2}{*}{\textbf{Language}}} & \multicolumn{2}{c|}{\textbf{Infusing Prompt}} & \multicolumn{2}{c|}{\textbf{Previous system}} \\
    \cline{2-5}
    & \textbf{EM} & \textbf{EX}  & \textbf{EM} & \textbf{EX} \\
    \hline
    Portuguese & 0.752 & 0.804 & 0.687 & - \\
    French & 0.749 & 0.765 & 0.698 & - \\ 
    \hline
   
\end{tabular}
\caption{The best models trained using our method compared to previous state of art results.}
\label{tablecomparison}
\end{table}

We also note   that, comparing the performance of a model to its counterpart in the parallel group of training epochs (keeping the same architecture and for the same language but training for a different number of epochs), the models infused with syntactic or semantic information trained for 32 epochs had values for both metrics near the values from their counterparts trained for 128 epochs. This suggests an expressive gain in training efficiency, allowing results near the values one would get after training the model for 128 epochs only after 32 epochs of training with syntax or semantics. This presents a reduction in the time cost of training the model, as well as in the energy consumed for it.
 
We have also ran experiments using the original Spider dataset, in English.
We saw little improvement for some cases, and in other a little decrease in performance,
when prompts received syntactic and semantic
information. For instance, the model T5 3B trained with AMR achieved exact set match of 0.688 and execution accuracy of 0.784, and the model without information had 0.714 and 0.787 for the respective metrics, whereas the best BART model trained with English was presented to AMR + syntax during the fine-tuning, and the T5 with highest exact-set match was the one fine-tuned with the English phrase and its syntactic information infused, as is exhibited by table \ref{table:theTableenglish128}.
We conjecture that a large language model trained with a vast number of
textual data somehow internalizes all information that could be given through
linguistic analysis.\footnote{Intuitively, and perhaps a bit theatrically: 
if a person knows little about a language, then getting some exposure to
syntactic and semantics lessons may be of significant help; but if a person
has years of exposure to a language,  syntactic and semantic lessons
may not be as relevant for basic understanding of the language.}

\section{Conclusion}\label{section:Conclusion}

In this paper we have a concise message: 
Syntactic and semantic information
should be used together with LLMs in low resource scenarios! 
The former can improve the latter, at low cost,
and in fact linguistic analysis can mitigate the need for ever larger annotated datasets.
This sort of message has been pursued in previous work, but we feel that we have here
described strategies that are simpler to apply, and that seem to be more effective.
It should be emphasized that we do not
require changes in inner components of transformers (such as positional embeddings,
self attention layers, etc). In fact,  it is not even necessary to change the inputs
at inference time -- even though we have shown that adding syntactic and semantic
information at inference time, and not only during training, is rather valuable.
Our prompting strategies should be added to the growing number
of ``prompting patterns'' that have been collected recently,
from few-shot to chain of thought prompting. 

Our experiments show that syntactic and semantic infusion helps in NL2SQL from
four natural languages, often
in a smaller number of training epochs. The representation that made the best contribution
was the one based on AMR graphs, even though dependency trees are also valuable.
We close with two key points that our experiments indicate about our proposals.
\begin{itemize}
\item Effectiveness: the positive value of training transformers with 
syntactic and semantic information, at least for low resource scenarios, without   retraining the models (linguistic analysis led models
to attain higher performance than possible otherwise).
\item Efficiency: syntactic and semantic information do accelerate the training process,
enabling models to reach target performance levels in fewer training epochs.
\end{itemize}

Concerning future work, we suggest that other tasks in natural
language processing should be investigated as to whether syntactic
and semantic information lead to better performance.
Also, we suggest that other architectures
should be tested; for instance, decoder only models may
benefit from syntactic and semantic information as such
models often struggle when dealing with low-resource 
languages \cite{lai2023chatgptenglishcomprehensiveevaluation}, \cite{hangya-etal-2022-improving}, \cite{bang-etal-2023-multitask}. 
Another key (and complex) issue to study is whether 
large language models
learn, in some suitable way, to do linguistic analysis when
exposed to ``sufficiently large'' resources --- that is,
to understand whether there are regimes where explicit
addition of syntactic and
semantic information does not make a difference as such
information is already internally produced. 
Obtaining insight on this issue would help one 
better design systems by balancing the resources devoted
to a large language model against the level of linguistic
information extracted from input and infused into the 
language model.


\appendix

\section{Evaluating the Translated Spider}

As indicated in Section 4, we run an evaluation of the translated Portuguese
version of the Spider dataset, to obtain a sense of the ``worst-case'' effect
of the translation through Google Translation API done by 
\cite{jose2023multilingual}.

We hired an experienced English/Portuguese teacher to verify the quality of the translation of 300 pairs of requests. These 300 samples were randomly picked from the training data, and were  sampled so that their proportion kept the proportion of ``hardness scores'' found in the training  according to the Spider classification. The Spider training dataset has 1609 easy, 2377 medium, 1626 hard and 1388 extra hard queries. To respect this proportion, we picked 69 easy, 102 medium, 70 hard and 59 extra hard queries. 
For each query, we asked the professional translator to evaluate the translation in a score from 1 to 5. The criteria  for grading was based on five critical points: 
(1) Does the phrase maintain the original phrase's meaning (semantic)?
(2) Does the sentence respect the standard (formal) Portuguese?
(3) Does the translated phrase seems to be the most natural translation?
(4) Is the translated phrase adding content to the original query?
(5) Is any information missing from the original query in the translated sentence? 

After this process, we were able to get a grasp of the quality of the automatically translated and manually verified by \cite{jose2023multilingual} dataset and the errors that we might find in it. The scores from the translator are summarized by query difficulty in Table \ref{table:scoretranslation}.  

 \begin{table}[t]
\centering
\begin{tabular}{l|l}
    Query difficulty & Score  \\ \hline
    Easy & 4.28 \\
    Medium & 4.41 \\
    Hard & 4.43 \\
    Extra & 4.37 
\end{tabular}
\caption{Scores for each difficulty level for the sampled queries in Spider}
\label{table:scoretranslation}
\end{table}

The most serious mistakes by the automatic translation tool generated
a few incorrect terms in the translated query, and did not change the
query to the point that it corresponded to a different SQL instruction. 
We were thus confident that the translated datasets could be used safely
in our experiments. 
Of course, as pointed out by \cite{min-etal-2019-pilot} and \cite{bakshandaeva2022pauq},  a human translated dataset might 
lead to better performance still for all models, but our main interest
here is in the relative performance as we want to assess whether our
techniques add value (all tests in a single language use the same 
translated Spider dataset).

\bibliography{aaai25}

\end{document}